\relax
\documentclass[letterpaper]{article} 
\usepackage{aaai22}  
\usepackage{times}  
\usepackage{helvet}  
\usepackage{courier}  
\usepackage[hyphens]{url}  
\usepackage{graphicx} 
\usepackage{natbib}  
\usepackage{caption} 
\DeclareCaptionStyle{ruled}{labelfont=normalfont,labelsep=colon,strut=off} 
\usepackage{algorithm}
\usepackage{algorithmic}
\usepackage{bbding}
\usepackage{colortbl}
\usepackage{listings}
\usepackage[switch]{lineno}
\usepackage[dvipsnames]{xcolor}
\usepackage[breaklinks=true, bookmarks=false, colorlinks=true, citecolor=RoyalBlue]{hyperref}
\usepackage{xspace}
\usepackage{multirow}
\usepackage{amssymb}

\usepackage{etoolbox}
\makeatletter
\AfterEndEnvironment{algorithm}{\let\@algcomment\relax}
\AtEndEnvironment{algorithm}{\kern2pt\hrule\relax\vskip3pt\@algcomment}
\let\@algcomment\relax
\newcommand\algcomment[1]{\def\@algcomment{\footnotesize#1}}
\renewcommand\fs@ruled{\def\@fs@cfont{\bfseries}\let\@fs@capt\floatc@ruled
  \def\@fs@pre{\hrule height.8pt depth0pt \kern2pt}%
  \def\@fs@post{}%
  \def\@fs@mid{\kern2pt\hrule\kern2pt}%
  \let\@fs@iftopcapt\iftrue}
\makeatother

\newcommand{\model}{\mbox{\sc{LinMapper}}\xspace}

\definecolor{mygray}{gray}{0.93}

\usepackage{newfloat}

\title{What Makes for Hierarchical Vision Transformer?}

\author{\textmd{Yuxin Fang$^{1}$\thanks{Preprint.
A part of this work was done when Yuxin Fang was interning at Horizon Robotics mentored by Rui Wu.}, 
\ \  Xinggang Wang$^{1}$\thanks{Xinggang Wang is the corresponding author.}, 
\ \ Rui Wu$^{2}$, 
\ \ Wenyu Liu$^{1}$ \\ 
{\normalsize $^1$School of EIC, Huazhong University of Science \& Technology \ \ \ 
$^2$ Horizon Robotics} \\
}
{\small \tt{\{yxf, xgwang\}@hust.edu.cn}}
}

\begin{document}

\maketitle

\begin{abstract}
Recent studies indicate that hierarchical Vision Transformer with a macro architecture of interleaved non-overlapped window-based self-attention \& shifted-window operation is able to achieve state-of-the-art performance in various visual recognition tasks, and challenges the ubiquitous convolutional neural networks (CNNs) using densely slid kernels. Most follow-up works attempt to replace the shifted-window operation with other kinds of cross-window communication paradigms, while treating self-attention as the de-facto standard for window-based information aggregation. In this manuscript, we question whether self-attention is the only choice for hierarchical Vision Transformer to attain strong performance, and the effects of different kinds of cross-window communication. To this end, we replace self-attention layers with embarrassingly simple linear mapping layers, and the resulting proof-of-concept architecture termed as \model can achieve very strong performance in ImageNet-$1k$ image recognition. Moreover, we find that \model is able to better leverage the pre-trained representations from image recognition and demonstrates excellent transfer learning properties on downstream dense prediction tasks such as object detection and instance segmentation. We also experiment with other alternatives to self-attention for content aggregation inside each non-overlapped window under different cross-window communication approaches, which all give similar competitive results. Our study reveals that the \textbf{macro architecture} of Swin model families, other than specific aggregation layers or specific means of cross-window communication, may be more responsible for its strong performance and is the real challenger to the ubiquitous CNN's dense sliding window paradigm. Code and models will be publicly available to facilitate future research.
\end{abstract}

\section{Introduction}

Recently, the impregnable position of convolutional neural networks (CNNs) in computer vision seems to be weakened by the emerging hierarchical Vision Transformer families~\cite{halonet, pvt, swin, MViT}.
As one representative, Swin Transformer~\cite{swin} and its variants (\emph{e.g.},~\citeauthor{shuffle, msg}) with an interleaved non-overlapped window-based self-attention \& cross-window token mixing paradigm are able to achieve state-of-the-art performance in image recognition, and demonstrate excellent transferability on various downstream computer vision tasks such as object detection and scene parsing.
Therefore it is meaningful to conduct an in-depth study on this new architecture family, and analyze what makes it so strong.

\begin{table}[t!]
    \centering
    \small
    \setlength{\tabcolsep}{9.8 pt}
    \begin{tabular}{l||c|c|c|c}

     & \cellcolor{red!5}MHSA
     & \cellcolor{red!5}Linear
     & \cellcolor{red!5}DW Linear 
     & \cellcolor{red!5}MLP
     \\
     
     \hline
     \hline
     
     \cellcolor{blue!5}Shift
     & $80.5$
     & $79.7$
     & $79.8$
     & $79.9$
     \\
     
     \cellcolor{blue!5}Shuffle
     & $80.6$
     & $79.6$
     & $79.6$
     & $79.8$
     \\
     
     \cellcolor{blue!5}MSG
     & $80.4$
     & $79.5$
     & $79.7$
     & $79.7$
     \\

    \end{tabular}
    \caption{Study of three representative hierarchical Vision Transformer with different information aggregation layers on ImageNet-$1k$~\cite{imagenet-1k}.
    All models are trained with \textbf{200} epochs schedule using the optimization scheme from~\cite{swin}.
    ``Shift'', ``Shuffle'', and ``MSG'' refer to the cross-window communication scheme in~\citeauthor{swin},~\citeauthor{shuffle}, and~\citeauthor{msg}, respectively.
    Models using MHSA layers keep the same configurations as their original implantations.
    Other models are tuned and selected with their \textit{best} performance consuming $\sim 4.5$G FLOPs budgets and parameters ranging from $24$M to $29$M with only model depth \& model width adaptations.
    We remove all densely slid conv-layers in the network stem and each block of~\cite{shuffle} for a clearer study of different aggregation layers in non-overlapped windows. 
    Overall, these results motivate us to focus more on the macro architecture design than specific aggregation layers or specific means of cross-window communication.}
    \label{tab: diff token talking & ops results.}
\end{table}

Methodologically, Swin Transformer first partitions feature maps to a series of non-overlapped local windows, and uses multi-head self-attention (MHSA) layers to aggregate information in each window individually.
Then, instead of using a dense sliding window paradigm for cross-window token mixing like CNNs~\cite{AlexNet, VGG, ResNet}, Swin Transformer proposes to shift windows between consecutive layers.
By alternating these two operations, each token is able to interact with all other tokens, and the overall architecture can obtain very strong capacities.

Most successors of Swin Transformer mainly focus on replacing the shifted-window operation with other kinds of cross-window communication, such as the spatial token shuffle operation from~\citeauthor{shuffle}, and the messenger tokens exchange proposed in~\citeauthor{msg}.
While the use of window-based MHSA is usually taken for granted and treated as the de-facto standard for token fusion within each local window.

In this manuscript, we ponder the question: ``What Makes for Hierarchical Vision Transformer?''.
Specifically, we investigate whether MHSA is the only choice to aggregate information for the Swin model family, and the effects of different cross-window communication schemes such as spatial token shuffle and auxiliary messager tokens exchange.
Previous practice from computer vision suggests that MHSA is good at capture dense and long-range contextual information~\cite{non-local, ccnet} in scene parsing tasks~\cite{tu2005image, tighe2014scene}.
Intuitively, it is somewhat \textbf{too aggressive} to use MHSA to model contextual relation inside a local window with only $7 \times 7 = 49$ spatial tokens.

This motivates us to replace MHSA layers with linear mapping, one of the most common \& simplest components in neural architecture design, in three representative hierarchical Vision Transformer instantiations, \emph{i.e.}, Swin Transformer~\cite{swin}, Shuffle Transformer~\cite{shuffle}, and MSG Transformer~\cite{msg}.
The resulting proof-of-concept model is termed as \model.
We find that \model with embarrassingly simple linear mapping layers is sufficient for local content aggregation, and is able to achieve very competitive performance in ImageNet-$1k$ image recognition benchmark~\cite{imagenet-1k}.
Moreover, \model can better leverage the pre-trained representations from image recognition and demonstrate excellent transferability in downstream dense prediction tasks such as object detection \& instance segmentation.

Furthermore, we experiment with other variants for information aggregation inside each local window, \emph{e.g.}, depth-wise linear mapping with separable weights (denoted as ``DW Linear'' in Tab.~\ref{tab: diff token talking & ops results.}), and multi-layer perceptrons with intermediate activation functions (denoted as ``MLP'' in Tab.~\ref{tab: diff token talking & ops results.}).
We also conduct a study on different cross-window communication approaches such as spatial token shuffle (denoted as ``Shuffle'' in Tab.~\ref{tab: diff token talking & ops results.}) and auxiliary messager tokens exchange (denoted as ``MSG'' in Tab.~\ref{tab: diff token talking & ops results.}).
As shown in Tab.~\ref{tab: diff token talking & ops results.}, we find different content aggregation layers all give similar competitive results under the same cross-window communication approaches, and vice versa: different cross-window communication methods also achieve similar strong performance under the same content aggregation layer.

Based on the available evidence, we hypothesize that the \textbf{macro architecture} of Swin model families, \emph{i.e.}, interleaved non-overlapped window-based token mixing \& cross-window communications, other than specific aggregation layers such as MHSA or specific means of cross-window communication such as shifted-window or spatial shuffle, may be more responsible for their strong performance, and is the real challenger to the ubiquitous CNN's dense sliding window paradigm.

Please note that this manuscript is \textbf{not} an attempt to show that simple linear or MLP layers are superior to MHSA.
On the contrary, we find MHSA is better than linear mapping \& MLP in terms of accuracy with even fewer budgets (see Tab.~\ref{tab: wider or deeper.}).
Our goal is to abstract away from specific aggregation layers as well as cross-window communication approaches, and highlight the importance and contribution of the macro architecture of the Swin Transformer family.
We hope our work can encourage the community to rethink the role of attention in neural architecture design, and shed a little light on future studies of general visual representation learning.

\section{Background and Related Work}

Highly mature and robust training recipes~\cite{deit} enable standard Transformer architecture~\cite{Transformer, vit} directly inherited from natural language processing to attain excellent performance in the image recognition task even with limited data \& model sizes~\cite{T2T, TNT}.
Standard Transformer models sequence-to-sequence relationship in a pair-wise manner with minimal prior via global scaled dot-product multi-head self-attention (MHSA), which scales quadratically with the sequence length.
Therefore standard Transformer suffers from the scaling problem in spatial dimensions and fails to process high-resolution inputs with varying sizes in computer vision downstream tasks.

To more efficiently apply Vision Transformers to other downstream tasks in computer vision such as object detection, instance segmentation, and scene parsing, three key issues need to be solved:
(1) involving hierarchical architectures to establish multi-scale feature representations for better handling of large variations in scales.
(2) reducing memory \& computation costs from global MHSA to efficiently process high-resolution inputs with varying token lengths, and 
(3) introducing appropriate inductive biases \& prior knowledge of the target task for better performance.

To mitigate the aforementioned issues, ~\citeauthor{pvt, PiT, MViT} process features with multi-resolution stages using spatial pooling operations instead of in a columnar manner.
~\citeauthor{halonet, swin} further propose to compute MHSA in weakly-overlapped or non-overlapped local windows.
After that, many follow-up hierarchical local window-based Vision Transformers emerge and challenge the hegemonic position of CNN in computer vision~\cite{shuffle, msg, NesT}.

To demystify the relation between CNN and hierarchical Vision Transformer,~\citeauthor{DWNet} study the inhomogeneous depth-wise convolution under modern training \& optimization recipe from~\citeauthor{swin}, and demonstrates that CNN can achieve similar competitive performance compared with the Swin Transformer family in various vision tasks.
Furthermore, ~\citeauthor{coat} and ~\citeauthor{coatnet} study the convolution-attention hybrid architecture. Combining the strengths from both camps, convolution-attention hybrid architecture can achieve state-of-the-art performance under different resource constraints across various datasets.

Previous studies show that simple multi-layer perceptrons (MLPs) architectures are competitive with CNNs in digit recognition~\cite{cirecsan2012deep, simard2003best}, keyword spotting~\cite{chatelain2006extraction} and handwritting recognition~\cite{bluche2015deep}.
Recently, a series of works~\cite{mlp-mixer, resmlp, thu-mlp} revisit the architecture based exclusively on columnar structured MLPs in image recognition tasks under modern training and transfer learning recipes. 

As cursorily summarized in Tab.~\ref{tab: missing piece.}, there are still two ``missing pieces'' remain, \textit{i.e.}, the CNN with columnar architectures, and the MLP with hierarchical architectures.
~\citeauthor{DWNet} touches the former topic with the columnar architecture proposed in~\citeauthor{volo}.
This manuscript conducts a primitive study to the latter one: a straightforward, simple, yet must-know model in computer vision.
We argue it is inevitable to investigate the potential of hierarchical linear mapping \& MLP structures now, and we hope the proposed proof-of-concept \model model can encourage the community to rethink the role between macro model design methodologies and specific network building blocks.

\begin{table}[t!]
    \centering
    \small
    \setlength{\tabcolsep}{4.2 pt}
    \begin{tabular}{l||c|c|c}

     & \cellcolor{red!5}Conv
     & \cellcolor{red!5}MHSA
     & \cellcolor{red!5}Linear \& MLP
     \\
     
     \hline
     \hline
     
     \cellcolor{blue!5}Columnar
     & ``Missing Piece''
     & \checkmark
     & \checkmark
     \\
     
     \cellcolor{blue!5}Hierarchical
     & \checkmark
     & \checkmark
     & \textbf{``Missing Piece''}
     \\

    \end{tabular}
    \caption{A cursory summary of macro architectures (\emph{Column}) and specific  aggregation layers (\emph{Row}). 
    ``\checkmark'' means the architecture configuration has been comprehensively studied, while there are still two ``missing pieces'' remain. This manuscript gives a primitively study to hierarchical architecture with simple linear or MLP layers.}
    \label{tab: missing piece.}
\end{table}

\section{What Makes for Hierarchical \\ Vision Transformer?}

We first briefly review the Swin Transformer family in Sec.~\ref{mth: swin}, and then introduce the proposed \model in Sec.~\ref{mth: linmap}.

\subsection{The Swin Architecture Family}
\label{mth: swin}
Methodologically, Swin Transformer~\cite{swin} processes high-resolution input hierarchically using multi-head self-attention (MHSA) within non-overlapped local windows.
Specifically, MHSA is used as the aggregation layer to fuse content information of spatial tokens inside each window.
Since the non-overlapped partition scheme lacks connection across windows, Swin Transformer proposes to use shifted-window operations between every two successive window-based MHSA layers to encourage cross-window communications.
Hierarchical architecture design is also adopted to produce multi-resolution representations for better handling of large scale \& size variations in visual entities.

Most successors of Swin Transformer mainly focus on replacing shifted-window operations with other kinds of cross-window communications such as spatial token shuffle~\cite{shuffle} or information exchange based on auxiliary messager tokens~\cite{msg}, while keeping other components unchanged.

Overall, Swin Transformer and its variants all adopt the MHSA as the \textbf{aggregation layer} for spatial token fusion, and use non-overlapped window-based token mixing \& cross-window communications in an alternating fashion with hierarchical representations as the \textbf{macro architecture}.
We refer readers to ~\citeauthor{swin, shuffle, msg} for more details of the Swin Transformer family architectures investigated in this manuscript.

\begin{algorithm}[t!]
\caption{ \ \ \model Pseudocode.}
\label{alg: code}
\definecolor{codeblue}{rgb}{0.25,0.5,0.5}
\definecolor{codekw}{rgb}{0.85, 0.18, 0.50}
\lstset{
  backgroundcolor=\color{white},
  basicstyle=\fontsize{8.0pt}{8.0pt}\ttfamily\selectfont,
  columns=fullflexible,
  breaklines=true,
  captionpos=b,
  commentstyle=\fontsize{8.0pt}{8.0pt}\color{codeblue},
  keywordstyle=\fontsize{8.0pt}{8.0pt}\color{codekw},
}
\begin{lstlisting}[language=python]
# B: num_windows, C: channel
# ws: window size, gs: group size
# t(dim1, dim2): # transpose dim1 & dim2

lin_map_h = Linear(ws*gs, ws*gs)
lin_map_w = Linear(ws*gs, ws*gs)
proj = PointWiseConv(C, C)

#  LinMapper
# x: input features with shape of (B, C, ws*ws)
def LinMapper(x): 
    
    # Height dim linear mapping for each window
    hf = x.view(B, C//gs, gs*ws, ws)
    hf = lin_map_h(hf.t(-1, -2)).t(-1, -2)
    hf = hf.view(B, C, ws*ws)
    
    # Width dim linear mapping for each window
    wf = x.view(B, C//gs, ws, gs*ws)
    wf = lin_map_w(wf)
    wf = wf.view(B, C, ws*ws)
    
return proj(hf + wf)
\end{lstlisting}
\end{algorithm}

\subsection{A Proof-of-concept Model: \model}
\label{mth: linmap}
Despite being greatly successful in various tasks, we question whether MHSA is the only choice to aggregate information for the Swin Transformer family.
To this end, we attempt to use linear mapping, one of the simplest components in neural architecture design, as a \textbf{touchstone \& probe} to reveal that the macro architecture (interleaved non-overlapped window-based token mixing \& cross-window communications) seems to be more responsible for Swin model families' strong performance other than specific aggregation layers such as MHSA.

\begin{table*}[t!]
    \centering
    \small
    \setlength{\tabcolsep}{7.575 pt}
    \begin{tabular}{l||cc|ccc||c}
    \hline
    
    \hline

    \hline
    
    \rowcolor{red!5}
     Method
     & Model Width
     & Model Depth
     & \#Params. (M)
     & FLOPs (G)
     & Throughput (Img/s)
     & Top-$1$ Acc.
     \\
     
     \hline
     \hline

     
     \cellcolor{blue!5}\textbf{Swin} \model-Tiny
     & $64$
     & $\{2, 4, 22, 4\}$
     & $24.6$
     & $4.0$
     & $320$
     & $80.5$
     \\
     
     
     
     \cellcolor{blue!5}\textbf{Shuffle} \model-Tiny
     & $64$
     & $\{2, 4, 22, 4\}$
     & $24.6$
     & $4.0$
     & $328$
     & $80.6$
     \\
     
     
     
     \cellcolor{blue!5}\textbf{MSG} \model-Tiny
     & $64$
     & $\{2, 4, 22, 4\}$
     & $30.6$
     & $4.5$
     & $380$
     & $80.4$
     \\
     
     \hline
     
     \hline
     
     \hline
    \end{tabular}
    \vspace{-0.8em}
    \caption{\textbf{Results of different \model-Tiny variants} on ImageNet-$1k$ image recognition benchmark.}
    \label{tab: main results.}
\end{table*}

\begin{table*}[t!]
    \centering
    \small
    \setlength{\tabcolsep}{7.8pt}
    \begin{tabular}{l||cc|ccc||c}
    \hline
    
    \hline

    \hline
    
    \rowcolor{red!5}
     Method
     & Model Width
     & Model Depth
     & \#Params. (M)
     & FLOPs (G)
     & Throughput (Img/s)
     & Top-$1$ Acc.
     \\
     
     \hline
     \hline
     
     \cellcolor{blue!5}Swin \model-Tiny
     & \textcolor{blue}{$64$}
     & $\{2, 4, 22, 4\}$
     & $24.6$
     & $4.0$
     & $320$
     & $80.5$
     \\
     
     \cellcolor{blue!5}Swin \model-Small
     & \textcolor{blue}{$96$}
     & $\{2, 4, 22, 4\}$
     & $54.9$
     & $8.9$
     & $201$
     & $82.0$
     \\
     
     \cellcolor{blue!5}Swin \model-Base
     & \textcolor{blue}{$128$}
     & $\{2, 4, 22, 4\}$
     & $97.3$
     & $15.9$
     & $143$
     & $82.5$
     \\
     
     \hline
     
     \hline
     
     \hline
    \end{tabular}
    \vspace{-0.8em}
    \caption{\textbf{Scaling the tiny-sized \model model}. We use the shifted-window operation~\cite{swin} as the default cross-window communication scheme.}
    \label{tab: scaling linmap.}
\end{table*}

We choose three representative and publicly available instantiations from the Swin model family, \textit{i.e.}, Swin Transformer~\cite{swin}, Shuffle Transformer~\cite{shuffle}, and MSG Transformer~\cite{msg}.
We directly replace their window-based MHSA layers with \model layers described in Algorithm~\ref{alg: code}, and align other components and configurations with Swin Transformer~\cite{swin}.

Specifically, \model layer takes partitioned image features with shape \texttt{(}$\mathtt{B, C, ws^2}$\texttt{)} as inputs, where $\mathtt{B}$ is the total number of partitioned windows, $\mathtt{C}$ is the number of channels and $\mathtt{ws}$ is the window size.
Since directly performing dense linear mapping from flattened inputs with shape \texttt{(}$\mathtt{B, C \times ws^2}$\texttt{)} to outputs with the same shape \texttt{(}$\mathtt{B, C \times ws^2}$\texttt{)} is computationally infeasible, we divide the input tensor to several groups where each group has $\mathtt{ws^2}$ tokens with $\mathtt{gs}$ channels (\textit{i.e.}, the group size is $\mathtt{gs}$).
To further reduce FLOPs and parameters budgets, linear mappings are performed along tokens' height and width dimensions separately, which is similar to the merit of~\citeauthor{ccnet} and~\citeauthor{axial}.
Therefore, the number of inputs for linear mapping is only $\mathtt{gs \times ws}$.
Finally, the transformed representations along two axes are fused via element-wise addition followed by a point-wise linear projection layer.

In our default instantiation, the weights and biases of $\mathtt{lin\_map\_h}$\texttt{()} and $\mathtt{lin\_map\_w}$\texttt{()} in Alg.~\ref{alg: code} are \textit{shared} across different groups.
Linear mappings with separate parameters for different groups are denoted as depth-wise linear mapping (``DW Linear'' in Tab.~\ref{tab: diff token talking & ops results.} and Tab.~\ref{tab: linear and dw linear.}) in this manuscript.
Our controlled experiments demonstrate that using separate weight cannot bring further improvements in linear mapping given similar model sizes, which echos the observation in previous multi-head \textit{v.s.} single-head self-attention studies~\cite{liu2021multi, michel2019sixteen}.

To some extent, \model is one of the simplest possible instantiations of architectures with interleaved non-overlapped window-based token mixing \& cross-window communications scheme.
Despite being simple, the \model layer is more lightweight than the MHSA layer.
Therefore the lack in capacity compared with MHSA can be compensated by deeper or wider architecture design given similar FLOPs \& parameters budgets, and the resulting architecture is still able to achieve competitive performance (Sec.~\ref{sec: main results.}).

It is noteworthy that the proposed \model layer enables a fully linear mapping \& MLP architecture to directly process high-resolution input images with arbitrary shapes.
This property allows MLP-like architectures to be easily transferred to different computer vision downstream tasks, which is lack in previous studies~\cite{mlp-mixer, resmlp, gmlp}.
In Tab.~\ref{tab: od & is.} we demonstrate that the transfer learning performance of \model is on a par with Swin Transformer in object detection and instance segmentation even with relatively worse supervised pre-trained representations on ImageNet-$1k$~\cite{imagenet-1k}.

Overall, the merit of \model is to abstract away from specific aggregation layers as well as cross-window communication schemes, and highlights the importance of the macro architecture, which seems overlooked in previous research on hierarchical Vision Transformer.

\section{Experiments}

We first give the general experimental setup in Sec.~\ref{sec: setup.}, and then report the pre-training, scaling, and transfer learning performance of \model in Sec.~\ref{sec: main results.}.
The model analysis and ablation study are finally conducted in Sec.~\ref{sec: analysis and ablation.}.

\subsection{Setup}
\label{sec: setup.}

\paragraph{Pre-train Settings.}
The experiments are conducted on the public available codebase of~\citeauthor{swin, shuffle, msg} and the $\mathtt{timm}$ library~\cite{timm}.

During pre-training, all models are trained and evaluated on ImageNet-$1k$~\cite{imagenet-1k} benchmark following the setup in~\cite{swin, deit}.
We train models with $300$ epochs on ImageNet-$1k$ for main results.
For model analysis and ablation study, we study different content aggregation layers with model width and model depth\footnote{In this manuscript, we define \textit{model width} is the number of channels in the first stage of the network, and \textit{model depth} is the number of content aggregation layers in each stage of the network.} same as Swin Transformer-Tiny (\textit{i.e.}, model width $= 96$, model depth $= \{2, 2, 6, 2\}$) using $200$ epochs training schedule on ImageNet-$1k$ unless specified.

The input resolution is $224 \times 224$ and the window size is $7 \times 7$ for all \model models in all experiments.
For a clearer study of different aggregation layers in non-overlapped windows, we remove all densely slid conv-layers in the network stem and each block of~\citeauthor{shuffle} in this manuscript.

Model throughput data during inference are measured using a single Titan Xp GPU with batch size $= 64$ with input resolution $= 224 \times 224$.
Model FLOPs during inference are measured with batch size $= 1$ with input resolution $= 224 \times 224$.

\begin{table*}[t!]
    \centering
    \small
    \setlength{\tabcolsep}{16.0 pt}
    \begin{tabular}{l||ccc||c}
    \hline
    
    \hline

    \hline
    
    \rowcolor{red!5}
     Method
     & Input Resolution
     & \#Params. (M)
     & FLOPs (G)
     & Top-$1$ Acc.
     \\
     
     \hline
     \hline

     \cellcolor{blue!5}MLP-Mixer-B/16~\cite{mlp-mixer}
     & $224$
     & $59$
     & $12.7$
     & $77.3$
     \\

     \cellcolor{blue!5}ResMLP-24~\cite{resmlp}
     & $224$
     & $30$
     & $6.0$
     & $79.4$
     \\

     \cellcolor{blue!5}ResMLP-36~\cite{resmlp}
     & $224$
     & $45$
     & $8.9$
     & $79.7$
     \\
     
     \cellcolor{blue!5}gMLP-S~\cite{gmlp}
     & $224$
     & $20$
     & $4.5$
     & $79.6$
     \\

     \cellcolor{blue!5}GFNet-S~\cite{GFNet}
     & $224$
     & $25$
     & $4.5$
     & $80.0$
     \\

     \cellcolor{blue!5}S$^2$-MLP-wide~\cite{S2MLP}
     & $224$
     & $71$
     & $14.0$
     & $80.0$
     \\

     \cellcolor{blue!5}Swin-Mixer-T/D6~\cite{swin}
     & $256$
     & $23$
     & $4.0$
     & $79.7$
     \\

     \cellcolor{blue!5}Swin \model-Tiny (\textbf{Ours})
     & $224$
     & $25$
     & $4.0$
     & $\mathbf{80.5}$
     \\
     
     \hline
     
     \cellcolor{blue!5}GFNet-B~\cite{GFNet}
     & $224$
     & $43$
     & $7.9$
     & $80.7$
     \\

     \cellcolor{blue!5}S$^2$-MLP-deep~\cite{S2MLP}
     & $224$
     & $51$
     & $10.5$
     & $80.7$
     \\

     \cellcolor{blue!5}Swin-Mixer-B/D24~\cite{swin}
     & $256$
     & $61$
     & $10.4$
     & $81.3$
     \\
     
     \cellcolor{blue!5}gMLP-B~\cite{gmlp}
     & $224$
     & $73$
     & $15.8$
     & $81.6$
     \\
     
     \cellcolor{blue!5}Swin \model-Small (\textbf{Ours})
     & $224$
     & $55$
     & $8.9$
     & $\mathbf{82.0}$
     \\
     

     \hline
     
     \hline
     
     \hline
    \end{tabular}
    \vspace{-0.8em}
    \caption{\textbf{Comparisons with other MLP-like architectures} on ImageNet-$1k$ image recognition benchmark.}
    \label{tab: mlp results.}
\end{table*}

\begin{table*}[t!]
    \centering
    \small
    \setlength{\tabcolsep}{4.2pt}
    \begin{tabular}{l||cc||c||ccc|ccc}
    \hline
    
    \hline

    \hline
    
    \rowcolor{red!5}
     Method
     & \#Params. (M)
     & FLOPs (G)
     & Pre-train Top-$1$
     & AP$^{bb}$
     & AP$^{bb}_{50}$
     & AP$^{bb}_{75}$
     & AP$^{m}$
     & AP$^{m}_{50}$
     & AP$^{m}_{75}$
     \\
     
     \hline
     \hline

     \cellcolor{blue!5}Swin Transformer-Tiny~\cite{swin}
     & $48$
     & $267$
     & $81.3$
     & $43.7$
     & $66.6$
     & $47.7$
     & $39.8$
     & $63.3$
     & $42.7$
     \\
     
     \cellcolor{blue!5}Swin \model-Tiny (\textbf{Ours})
     & $44$
     & $253$
     & $80.5$
     & $43.8$
     & $66.2$
     & $47.8$
     & $39.6$
     & $62.9$
     & $42.4$
     \\

     \hline
     
     \hline
     
     \hline
    \end{tabular}
    \vspace{-0.8em}
    \caption{\textbf{Transfer learning performance} of \model-Tiny on MS-COCO object detection and instance segmentation benchmarks using the \textbf{Mask R-CNN} framework~\cite{MaskRCNN} with training \& testing configurations from~\citeauthor{swin}.}
    \label{tab: od & is.}
\end{table*}

\paragraph{Transfer Learning Settings.}
The experiments are conducted on the public available codebase of~\citeauthor{swin} and the $\mathtt{mmdetection}$ library~\cite{mmdetection}.

We study the transfer learning performance of ImageNet-$1k$ $300$-epoch supervised pre-trained \model models in the challenging MS-COCO~\cite{COCO} object detection and instance segmentation benchmarks using the Mask R-CNN~\cite{MaskRCNN} framework.
We fine-tune the pre-trained \model with standard $1 \times$ schedule~\cite{MaskRCNN} on MS-COCO $\mathtt{train}$ split and report the transfer learning results on MS-COCO $\mathtt{val}$ split following the training and testing configurations from~\citeauthor{swin}.

Model FLOPs during inference are measured with batch size $= 1$ and input resolution $= 1280 \times 800$.

\subsection{Main Results}
\label{sec: main results.}

\paragraph{Results of \model on ImageNet-1k.}
As shown in Tab.~\ref{tab: main results.}, given limited FLOPs and parameters budgets, tiny-sized \model models are able to achieve competitive performance on ImageNet-$1k$~\cite{imagenet-1k} image recognition benchmark with three different cross-window communication paradigm, \emph{i.e.}, shifted-window~\cite{swin}, spatial token shuffle~\cite{shuffle}, and auxiliary messager tokens exchange~\cite{msg}.

Along with the results in Tab.~\ref{tab: diff token talking & ops results.}, it is noteworthy that (1) shifted-window, spatial token shuffle, and auxiliary messager tokens exchange all give similar strong results under the same aggregation layer, and moreover, (2) different aggregation layers are all quite competitive under the same cross-window token mixing approach.

All these results support our proposal: the macro architecture of the Swin model family, other than specific aggregation layers or specific means of cross-window communication, may be more responsible for its strong performance.

\begin{table*}[t]
    \centering
    \small
    \setlength{\tabcolsep}{4.8pt}
    \begin{tabular}{l||ccccc||c}
    \hline
    
    \hline

    \hline
    
    \rowcolor{red!5}
     Method
     & Model Width
     & Model Depth
     & \#Params. (M)
     & FLOPs (G)
     & Throughput (Img/s)
     & Top-$1$
     \\
     
     \hline
     \hline

     \cellcolor{blue!5}Swin Transformer-Tiny~\cite{swin}
     & $96$
     & $\{2, 2, 6, 2\}$
     & $28.3$
     & $4.5$
     & $378$
     & $80.5$
     \\
     
     \hline
     
     \cellcolor{blue!5}Swin \model-Tiny (Baseline)
     & $96$
     & $\{2, 2, 6, 2\}$
     & $22.6$
     & $3.4$
     & $410$
     & $78.8$
     \\
     
     \cellcolor{blue!5}Swin \model-Tiny (Wide)
     & \textcolor{blue}{$112$}
     & $\{2, 2, 6, 2\}$
     & $30.6$
     & $4.6$
     & $343$
     & $79.8$
     \\
     
     \cellcolor{blue!5}Swin \model-Tiny (Deep)
     & $64$
     & \textcolor{blue}{$\{2, 4, 22, 4\}$}
     & $24.6$
     & $4.0$
     & $320$
     & $79.7$
     \\
     
     \hline
     
     \hline
     
     \hline
    \end{tabular}
    \vspace{-0.8em}
    \caption{\textbf{Wider \textit{v.s.} deeper} macro structure for \model.}
    \label{tab: wider or deeper.}
\end{table*}

\begin{table}[t]
    \centering
    \small
    \setlength{\tabcolsep}{3.pt}
    \begin{tabular}{c|l||ccc||c}
    \hline
    
    \hline

    \hline
    
    \rowcolor{red!5}
     \#Group 
     & Agg. Layer
     & \#Params.
     & FLOPs
     & Throughput
     & Top-$1$
     \\
     
     \hline
     \hline

     \cellcolor{blue!5}
     & \cellcolor{blue!5}Linear
     & $22.0$
     & 
     & 
     & $78.8$
     \\

     \cellcolor{blue!5}\multirow{-2}{*}{\shortstack[c]{$32$}}
     & \cellcolor{blue!5}DW Linear
     & $28.4$
     & \multirow{-2}{*}{\shortstack[c]{$3.3$}}
     & \multirow{-2}{*}{\shortstack[c]{$413$}}
     & $78.9$
     \\
     
     \hline
     
     \cellcolor{blue!5}
     & \cellcolor{blue!5}Linear
     & $21.8$
     & 
     &
     & $78.7$
     \\

     \cellcolor{blue!5}\multirow{-2}{*}{\shortstack[c]{$48$}}
     & \cellcolor{blue!5}DW Linear
     & $26.2$
     & \multirow{-2}{*}{\shortstack[c]{$3.3$}}
     & \multirow{-2}{*}{\shortstack[c]{$416$}}
     & $78.9$
     \\
     
     \hline
     
     \hline
     
     \hline
    \end{tabular}
    \vspace{-0.8em}
    \caption{\textbf{Linear \textit{v.s.} depth-wise linear} in \model.}
    \label{tab: linear and dw linear.}
\end{table}

\begin{table}[t]
    \centering
    \small
    \setlength{\tabcolsep}{4.8pt}
    \begin{tabular}{c||ccc||c}
    \hline
    
    \hline

    \hline
    
    \rowcolor{red!5}Agg. Layer
     & \#Params.
     & FLOPs
     & Throughput
     & Top-$1$
     \\
     
     \hline
     \hline
     
     \cellcolor{blue!5}MLP
     & $22.8$
     & $3.4$
     & $357$
     & $79.0$
     \\

     \cellcolor{blue!5}Linear Mapping
     & $22.0$
     & $3.3$
     & $413$
     & $78.8$
     \\
     
     \hline
     
     \hline
     
     \hline
    \end{tabular}
    \vspace{-0.8em}
    \caption{\textbf{Linear mapping \textit{v.s.} MLP} in \model.}
    \label{tab: linear and mlp.}
\end{table}

\paragraph{Scaling \model-Tiny.}
There is little literature available on the scaling properties of hierarchical MLP-like models.
Here we demonstrate that the tiny-sized \model model is \textbf{scalable}.

We choose Swin \model-Tiny as the model scaling start point.
Width scaling~\cite{wresnet, mobilenets} is adopted while the model depth and input resolution are kept unchanged.
The training and testing configurations are aligned with~\citeauthor{swin}.
Results in Tab.~\ref{tab: scaling linmap.} show that both \model-Small \& \model-Base can be successfully optimized, converged, and consistently benefit from more computations and larger model size.

Finding appropriate model scaling laws tailored for \model as well as other MLP variants is non-trivial, since the model complexity of MLP-like architectures is a linear combination of two different parts: (1) the content aggregation layer part for depth-wise or group-wise spatial token mixing only, and (2) the feed-forward network part for point-wise or token-wise feature transformation only, which is different from the complexity of CNNs.
Consequently, previous successful practice in CNNs scaling cannot apply to \model in a principled way and can be only verified one-by-one experimentally.
We leave the study of sophisticated model scaling laws on \model for future work.

\paragraph{Comparisons with Other MLP Variants.}
We summarize some recently proposed MLP-like architectures in Tab.~\ref{tab: mlp results.}.
The proposed \model demonstrates superior performance with fewer FLOPs and parameters.
The window partitioning operation and hierarchical representations introduce $2\mathrm{D}$ locality bias and invariance to \model. 
Therefore it is not a surprise that \model is more efficient and competitive than global \& columnar MLP architectures such as MLP-Mixer and ResMLP by leveraging these design priors.

Swin-Mixer is a recently proposed MLP-like architecture with hierarchical representations based on the Swin Transformer family.
We demonstrate that both the tiny-sized and small-sized \model can outperform the corresponding Swin-Mixer counterparts even with fewer computation budgets.

\paragraph{Transfer Learning Performance of \model.}
There is little literature available on the transferability of MLP-like architecture to downstream dense prediction tasks such as object detection and instance segmentation.
Most available MLP variants using global kernels for spatial token mixing, therefore their spatial kernel sizes are fixed and highly correlated to the input resolution during pre-training, which largely limits the applications on high-resolution inputs with varying shapes and sizes during transfer learning.

The proposed \model can naturally overcome this issue via window-based spatial token mixing inherited from the Swin model family.
\model is able to directly process arbitrary resolutions zero-padded to be divisible by the window size.

Results in Tab.~\ref{tab: od & is.} demonstrate that the transfer learning performance of \model is on a par with Swin Transformer in MS-COCO object detection and instance segmentation benchmarks even using relatively worse supervised pre-trained representations on ImageNet-$1k$.
These promising results indicate that \model could be able to better leverage the pre-trained representations for transfer learning.

\subsection{\model Analysis and Ablation Study}
\label{sec: analysis and ablation.}

In this section, we study the impact of model width \& model depth configurations, weight sharing properties of linear mapping layers, as well as the number of groups.
Overall, we conclude that \model is quite robust to different model choices and configurations thanks to the strong macro architecture.

\paragraph{Going Wider or Deeper?}
We study the model width and model depth configurations for the tiny-sized \model model.
A high-performance tiny-sized model can also be served as a promising start point for model scaling.

Since a single \model layer is much lighter than a single window-based MHSA layer from~\citeauthor{swin} in terms of both parameters and computations, we need to adjust the model width and depth of \model to align with the budgets for Swin Transformer-Tiny.
The resulting models are termed as Swin \model-Tiny (Wide) and Swin \model-Tiny (Deep).

As shown in Tab.~\ref{tab: wider or deeper.}, the wider model seems to be more speed friendly while the deeper model is a lot more parameters \& FLOPs efficient.
We choose Swin \model-Tiny (Deep) as our default tiny-sized \model model instantiation.

\paragraph{Linear or Depth-wise Linear?}
In this manuscript, depth-wise linear (DW Linear) layers refer to linear mapping with separate or non-shared weights \& biases for each group.
Therefore the model parameters increase while the theoretical FLOPs are kept unchanged when using DW linear layers instead of shared linear layers.
As shown in Tab.~\ref{tab: linear and dw linear.}, DW linear layers bring no significant improvement, which echos the observation in previous multi-head \textit{v.s.} single-head self-attention studies~\cite{liu2021multi, michel2019sixteen} to some extent.
Therefore we choose to share linear weights \& biases across different groups as our default instantiation for better parameters efficiency.

\paragraph{Linear Mapping or MLP?}
In Tab.~\ref{tab: linear and mlp.}, we investigate the impact of using simple linear mapping layers and more sophisticated MLP layers for content aggravation inside each non-overlapped window.
The results suggest that using MLPs with intermediate $\mathtt{GELU}$\texttt{()} activation functions~\cite{gelu} cannot bring further improvements.

This observation in our hierarchical \model is somewhat in line with the findings from the columnar counterpart~\cite{resmlp}, where ResMLP models also adopt simple patch-to-patch (or token-to-token) linear transformations instead of MLP layers in MLP-Mixer~\cite{mlp-mixer} for cross-patch communications.

\paragraph{Number of Groups (\#Group) for Linear Layers.}
In Tab.~\ref{tab: group size.}, we study the impact of different numbers of groups (\#Groups) in linear layers.
The weights \& biases of linear layers are shared across groups.
We find setting \#Groups too large or too small is harmful to performance, while other choices yield similar results.
In this manuscript, \#Groups $=32$ (\textit{i.e.}, $\mathtt{gs} = 3 \times 2^{({\#}\mathtt{stage} - 1)}$ in Alg.~\ref{alg: code}) is chosen as the default configuration for all-sized \model models.

\begin{table}[t]
    \centering
    \small
    \setlength{\tabcolsep}{1.8pt}
    \begin{tabular}{c|r||ccc||c}
    \hline
    
    \hline

    \hline
    
    \rowcolor{red!5}
     \#Group 
     & $\mathtt{gs}$ in Alg.~\ref{alg: code} \ \ \
     & \#Params. 
     & FLOPs
     & Throughput
     & Top-$1$
     \\
     
     \hline
     \hline

     \cellcolor{blue!5}$96$
     & \cellcolor{blue!5}$1*2^{({\#}\mathtt{stage}-1)}$
     & $21.8$
     & $3.3$
     & $413$
     & $78.4$
     \\

     \cellcolor{blue!5}$48$
     & \cellcolor{blue!5}$2*2^{({\#}\mathtt{stage}-1)}$
     & $21.8$
     & $3.3$
     & $416$
     & $78.7$
     \\
     
     \cellcolor{blue!5}$32$
     & \cellcolor{blue!5}$3*2^{({\#}\mathtt{stage}-1)}$
     & $22.0$
     & $3.3$
     & $413$
     & $78.8$
     \\
     
     \cellcolor{blue!5}$16$
     & \cellcolor{blue!5}$6*2^{({\#}\mathtt{stage}-1)}$
     & $22.6$
     & $3.4$
     & $410$
     & $78.8$
     \\
     
     \cellcolor{blue!5}$8$
     & \cellcolor{blue!5}$12*2^{({\#}\mathtt{stage}-1)}$
     & $25.0$
     & $3.4$
     & $420$
     & $78.6$
     \\
     
     \hline
     
     \hline
     
     \hline
    \end{tabular}
    \vspace{-0.8em}
    \caption{\textbf{Number of groups} in \model.}
    \label{tab: group size.}
\end{table}

\section{Conclusion}
In this manuscript, we raise a crucial question that seems overlooked in previous research: ``What Makes for Hierarchical Vision Transformer?'', and attempt to give an answer: the macro architecture design methodology may be more important than specific network layers and components.
To this end, a proof-of-concept model named \model with embarrassingly lightweight \& simple content aggregation layers is proposed as a touchstone or probe to support and validate our proposal.
As a hierarchical MLP-like architecture, \model along with its variants also enables a fully linear mapping \& MLP architecture to directly process high-resolution input images with arbitrary shapes, which makes it possible for investigating the transferability of linear mapping \& MLP architectures on various computer vision downstream tasks other than image recognition such as object detection and scene parsing.
We hope this manuscript can encourage the community to rethink the design of Vision Transformer, and shed a little light on future studies of general visual representation learning.

\bibliography{aaai22}

\end{document}